\documentclass[a4paper]{article}

\usepackage{algorithm}
\usepackage{algpseudocode}
\usepackage{amssymb,amsmath}
\usepackage{cite}
\usepackage{enumitem}
\usepackage[margin=1.5in]{geometry}
\usepackage[utf8]{inputenc}
\usepackage{microtype}
\usepackage{upquote}
\usepackage{epsfig}

\algnewcommand\algorithmicinput{\textbf{Input:}}
\algnewcommand\Input{\item[\algorithmicinput]}

\algnewcommand\algorithmicoutput{\textbf{Output:}}
\algnewcommand\Output{\item[\algorithmicoutput]}

\UseMicrotypeSet[protrusion]{basicmath}

\DeclareMathOperator{\atan2}{arctan2}

\newcommand{\eopra}[1]{$e\mathcal{OPRA}_#1$}
\newcommand{\opra}[1]{$\mathcal{OPRA}_#1$}
\title{Qualitative shape representation based on the
qualitative relative direction and distance calculus \eopra{m}}

\author{%
    {Christopher H. Dorr and Reinhard Moratz}\\\\%
    \small{NCGIA \& School of Computing and Information Science}\\%
    \small{University of Maine, ME, USA}}

\date{\today}

\begin{document}

    \maketitle

    \abstract%
    \label{abstract}

    This document serves as a brief technical report, detailing the processes used to represent and reconstruct simplified polygons using qualitative spatial descriptions, as defined by the \eopra{m} qualitative spatial calculus.

    \section{Overview}%
    \label{overview}

    Qualitative spatial reasoning (QSR) abstracts metrical details of the physical world and enables computers to make predictions about spatial relations even when precise quantitative information is unavailable \cite{cohn1997qualitative}. From a practical viewpoint qsr is an abstraction that summarizes similar quantitative states into one qualitative characterization. A complementary view from the cognitive perspective is that the qualitative method {\it compares} features within the object domain rather than by {\it measuring} them in terms of some artificial external scale \cite{freksa1992using}.  This is the reason why qualitative descriptions are quite natural for humans.

    The two main directions in QSR are topological reasoning about regions \cite{randell1992spatial}, \cite{renz1999complexity}, \cite{worboys2001integration} and positional reasoning about point configurations, like reasoning about orientation and distance \cite{freksa1992using}, \cite{clementini1997qualitative}, \cite{zimmermann1996qualitative}, \cite{isli1999qualitative}.

    There is also some work about using positional reasoning to describe the qualitative shape of 2D regions. These approaches represent qualitative shape by listing the relative positions of the adjacent vertices of polygons enumerating the outline of the polygon \cite{gottfried2002tripartite}. However, this work only makes very limited use of concepts of qualitative distance. Based on the recent work by Moratz and Wallgr{\"u}en \cite{moratz2014spatial} there is a candidate for a finer resolution positional QSR calculus called \eopra{m} which is suited to describe outlines of polygons on different levels of granularity.

    What is the motivation for using qualitative shape descriptions? Qualitative shape descriptions can implicitly act as a schema for measuring the similarity of shapes, which has the potential to be cognitively adequate. Then, shapes which are similar to each other would also be similar for a pattern recognition algorithm. There is substantial work in pattern recognition and computer vision dealing with shape similarity. Here with our approach to qualitative shape descriptions and shape similarity, the focus is on achieving a similarity which can be tested in human test subject empirical research to map to human intuitions about shape similarity. Such experiments became a standard tool in QSR \cite{klippel2013egenhofer}.

    To enable these experiments the qualitative shape representation must be somehow reversable. That means it must be possible to take the qualitative shape representation and generate prototypical specific shapes that match the abstract description. In previous work about QSR-based shape description it was only possible to take shapes and generate the abstract qsr-based description out it. It was not possible to take a qsr-based shape description and let an automatic algorithm generate a sample shape matching this descrption. Our work described in this report is the very first qsr-based shape description capable of generating prototypical shapes base on the abstract qsr-based representation.

    In the following report, we discuss the steps taken to reconstruct simple polylgons using \eopra{m} descriptions. For our purposes, polygons are defined as a simple closed polylines, or a non self-intersecting chain of points existing in the Cartesian plane $\mathbb{R}^2$. Inputs are converted into qualitative \eopra{m} descriptions, which are then reconstructed as polygons through a combination of state-space searching and constraint propagation.

    This report does not cover the process of creating the simplified input polylines. In short, noisy input polylines are simplified via the dce method presented in \cite{latecki1999polygon,barkowsky2000schematizing}.

    We show that given an appropriate level of granularity, the \eopra{m} calculus can be used to represent and reconstruct similar approximations of simple polygons.

    Results presented in this report are produced by a small set of Python programs developed to perform the deconstruction and reconstruction tasks. Roughly, the deconstruction and reconstruction is a three-step process:
    \begin{itemize}
        \item\label{overview-step1} compute the vertex-pairwise \eopra{m} direction and distance descriptions of the input polyline;
        \item\label{overview-step2} perform an initial reconstruction by simply ``tracing'' the qualitative hull description;
        \item\label{overview-step3} refine the results of \ref{overview-step2} via a greedy search.
    \end{itemize}

    The remainder of this report is structured as follows: in Section \ref{generating-qualitative-descriptions}, we begin with a brief primer on the \eopra{m} calculus, followed by a discussion on the details of generating qualitative descriptions from input polylines. In Section \ref{initial-reconstruction-and-refinement} we outline both the initial reconstruction and the subsequent refinement prodecures. In Section \ref{preliminary-results} we present some sample results. Lastly, in Section \ref{discussion-futwork}we offer some discussion on existing and future work.

    \section{Generating Qualitative Descriptions}%
    \label{generating-qualitative-descriptions}

    The goal of this section is to describe the process of generating an \eopra{m} qualitative representation from simple polyline data. First, a brief primer on the \eopra{m} calculus itself.

    \subsection{The \bf{\eopra{m}} Spatial Calculus}%
    \label{eopra-background}

    The \eopra{m} calculus \cite{moratz2014spatial} exists as an extension of the \opra{m} calculus \cite{moratz2006representing,mossakowski2012qualitative}. Under \opra{m}, points objects are augmented with a local reference direction to create ``oriented points,'' or \emph{o-points}. Using these local reference directions, along with a series of binary relations, one can express the notion of \emph{relative direction} between sets of o-points. The \opra{m} calculus also includes a granularity parameter $m$, which defines the resolution of these relative directions by partitioning the plane into $4m$ sectors. 

    The \eopra{m} calculus extends the \opra{m} calculus by adding the concept of \emph{relative distance} to o-points, yielding \emph{eo-points}. As with directions, local reference distances are attached to points enabling the comparison of relative distances. Similar to qualitative directions, qualitative spatial distances are defined by partitioning the plane into $2m$ sections, representing distances from 0 (none, overlapping) to \emph{same} (equal), to \emph{far} or infinity.

    \subsection{\bf{\eopra{m}} Descriptions of Polygons}%
    \label{eopra-descriptions}

    Qualitative \eopra{m} descriptions of polygons are comprised of three primary components: a granularity measure $m$, a pairwise set of qualitative directions, and a set of qualitative distances. 

    The first component, $m$, defines the granularity of our representation. Although \eopra{m} supports the use of different granularity measures for direction and distance, here we use one value for both measures. Once the value of $m$ is specified, it is used to create the pairwise qualitative direction matrix and to transform edge lengths into qualitative distances.

    \subsubsection{Qualitative Directions}%
    \label{qualitative-directions}

    To compute the pairwise qualitative direction matrix, one must first find the quantitative counter-clockwise turn angles between each pair of connected edges, or ordered triple of vertices. Given two edges $\mathbf{v}_1$ and $\mathbf{v}_2$ sharing a common vertex, the positive (counter-clockwise) turn angle $\beta$ is defined as:

    \begin{equation}
        \begin{aligned}
            \beta &= 
                \atan2{\left(\mathbf{v}_{y_2}, \mathbf{v}_{x_2}\right)} - 
                \atan2{\left(\mathbf{v}_{y_1}, \mathbf{v}_{x_1}\right)} \\
            \beta &= 
                \begin{cases}
                    \beta \qquad&\textbf{if}~\beta \ge 0 \\
                    \beta + 2\pi \qquad&\textbf{if}~\beta < 0
                \end{cases}
        \end{aligned}
    \end{equation}

    An outline of the process used to create all possible connected point triples and compute their turn angles is presented below. 
    
    %!TEX root = ../dce-eopra-report.tex
\noindent
\begin{minipage}{\linewidth}
\begin{algorithm}[H]
\caption{Generating Pairwise Turn Angles}\label{alg-pw-cct}
\begin{algorithmic}

    \Input Simple polygon of $n$ vertices, $p_0$ through $p_{n-1}$
        \Statex
        \State $O\gets$ empty $ n\times n$ matrix
        \For{$i\gets 0, \dotsc, n-1$}
            \For{$j\gets 0, \dotsc, n-1$}
                \If{$i\ne j$}
                    \Statex
                    \State $prev\gets i-1$
                    \Statex
                    \If{$i=0$}%
                    \Comment{if we are at the first vertex}
                        \State $prev\gets n-1$%
                        \Comment{then set \texttt{prev} to the last vertex}
                    \EndIf
                    \Statex
                    \State $\mathbf{v}_1\gets p_i - p_{prev}$
                    \State $\mathbf{v}_2\gets p_j - p_i$
                    \Statex
                    \State $O\left[i, j\right]\gets \mathrm{PositiveTurnAngleBetween}(\mathbf{v}_1, \mathbf{v}_2)$
                    \Statex
                \EndIf
            \EndFor
        \EndFor
        \State \textbf{return} $O$%
        \Comment{return populated turn matrix}
        \Statex
    \Output $O$, a $n\times n$ matrix of positive counter-clockwise turn angles

\end{algorithmic}
\end{algorithm}
\end{minipage}

    \vskip\baselineskip

    After all turn angles are computed, the next step is to convert them into relative directions. This is done simply by dividing each turn angle (reference direction) by the angular resolution specified by the \eopra{m} granularity measure $m$. The angular resolution of a given \eopra{m} representation is obtained by dividing the set of possible directions ($0-2\pi$ radians) into $4m$ partitions.

        \begin{equation}
            e\mathcal{OPRA}_m~\textrm{angular resolution} = 2\pi / 4m = \pi / 2m
        \end{equation}

        \begin{equation}\label{theta2oval}
            \textrm{turn angle}~\theta~\textrm{as}~e\mathcal{OPRA}_m~\textrm{direction} = \theta / \left(\pi / 2m\right) = \left(\theta\times2m\right) / \pi
        \end{equation}

        \begin{equation}\label{oval2theta}
            e\mathcal{OPRA}_m~\textrm{direction}~i~\textrm{as turn angle} = i\times\textrm{angular resolution}
        \end{equation}

        \vskip\baselineskip

    For example, given $m=2$, a reference direction of $\frac{\pi}{2}$ translates to an \eopra{2} direction partition of $\left(\frac{\pi}{2} \times 4\right) / \pi = 2$. When the result of eq. \ref{theta2oval} is not an exact integer, the direction partition is assigned as follows: if the integer part $i$ of the result is even, the angle is assigned to partition $i+1$. If the integer part is odd, the angle is assigned to partition $i$. 

        \begin{equation}
            e\mathcal{OPRA}_m~\textrm{direction interval}~i=
            \begin{cases}
                \lfloor i \rfloor+1&\textbf{if}~\lfloor i \rfloor~\textrm{is even}\\
                \lfloor i \rfloor&\textbf{if}~\lfloor i \rfloor~\textrm{is odd}
            \end{cases}
        \end{equation}

    \subsubsection{Qualitative Distances}%
    \label{qualitative-distances}
    
    The next step in creating our qualitative \eopra{m} description is to translate the absolute distances between adjacent vertices into qualitative reference distances. As with directions, \eopra{m} reference distances are computed by partitioning a plane into $2m$ sections, and assigning quantitative values a qualitative representation. Qualitative distances are computed by comparing each edge's length with the previous edge's length as a ``control'' length. Under \eopra{m}, a control length $\delta$ is used to turn a series of distance ratios into a series of qualitative distances.

    The sequence of distance ratios is defined by the granularity parameter $m$ such that there are $2m$ partitions: the first half of these ratios represent distances $\le\delta$ marked at even intervals of $0, \frac{1}{m}, \frac{2}{m}, \ldots, \frac{m-1}{m}, 1$. The second half of the ratios represent distances $>\delta$, and are reciprocals of the first half (reversed): $\frac{m}{m-1}, \frac{m}{m-2}, \ldots, m$. For example, given $m=4$, these ratios would be $\left(0, \frac{1}{4}, \frac{1}{2}, \frac{3}{4}, 1, \frac{4}{3}, 2, 4\right)$. Given these ratios, reference distances are calculated by multiplying each of the ratios by a control length to create quantitative distance markers.

    In our implementation, the qualitative distance ratios are created once given $m$, and applied over each edge of the input polygon. Using the previous edge as a control length $\delta$, each distance ratio is turned  into a quantitative distance. Once the pair of distances which bound the current edge length is identified, a qualitative distance is assigned as the average of the two bounding distances. For simplicity, we currently start this process by assigning the first edge's length to that of the input polygon's first edge.

    Unlike with the directions, this is not done pairwise, but instead only for the distances which represent the edges of the input polygon. 

    %!TEX root = ../dce-eopra-report.tex
\noindent
\begin{minipage}{\linewidth}
\begin{algorithm}[H]
\caption{Generating Qualitatve Distances}\label{alg-qual-dist}
\begin{algorithmic}

    \Input List of $n$ quantitative edge lengths $E_0, \dotsc, E_{n-1}$ from input polygon, list of qualitative distance ratios $D$
    \Statex
        \State $Q\gets$ empty list to hold qualitative distances
        \State $Q_0\gets E_0$%
        \Comment first edge length used as initial control length
        \Statex
        \For{$i\gets 0, \dotsc, n-1$}
            \State $control\gets Q_i$
            \State $target\gets E_{i+1}$
            \State $Q_{i+1}\gets \textsc{EdgeLengthToQSD}(D, control, target)$
        \EndFor
        \Statex
        \State \textbf{return} $Q$%
        \Comment{return list of qualtitative distances}
        \Statex
    \Output $Q$, a list of $n$ qualitative edge lengths

\end{algorithmic}
\end{algorithm}
\end{minipage}

    \vskip\baselineskip

    In the algorithm above, each consecutive pair of edges is used to convert an input edge length to a qualitative distance by comparing a $target$ edge length to a series of qualitative distances generated by multiplying each distance ratio by some $control$ distance. Given a control length $\delta$, a target length $t$ to be converted to a qualitative distance, and a set of qualitative distance ratios $D$, the process computing the qualitative distance goes like:

    %!TEX root = ../dce-eopra-report.tex
\noindent
\begin{minipage}{\linewidth}
\begin{algorithm}[H]
\caption{Converting Edge Lengths to Qualitative Distances}\label{alg-elen2qsd}
\begin{algorithmic}

    \Input List of $n$ qualitative distance ratios $D$, a control length $\delta$, and a target length $t$ to be converted
    \Statex
    \Procedure{EdgeLengthToQSD}{$D, \delta, t$}
        \Statex
        \State $M\gets D\times\delta$%
        \Comment multiply distance ratios by $\delta$
        \Statex
        \For{$i\gets 0, \dotsc, n-1$}
            \Statex
            \State $lower\gets M_i$
            \State $upper\gets M_{i+1}$
            \Statex
            \If{$lower \le t < upper$}
                \State \textbf{return} $\left(upper + lower\right)/2$%
            \EndIf
        \EndFor
        \Statex
    \EndProcedure

\end{algorithmic}
\end{algorithm}
\end{minipage}

    \vskip\baselineskip
    %More specifics on the \eopra{m} calculus can be found in \cite{moratz2014spatial}.

    At this point, we have completed generating qualitative descriptions of both the angles between vertices and the lengths of connected edges, and can proceed with the tasks involved in reconstruction.

    \section{Initial Reconstruction and Refinement}%
    \label{initial-reconstruction-and-refinement}

    The first step in the reconstruction process is to simply ``trace'' the hull of the \eopra{m} descriptions generated in \ref{qualitative-directions} and \ref{qualitative-distances}. To start, we turn the hull list of \eopra{m} direction intervals from eq. \ref{theta2oval} back into turn angles (by multiplying each value by the angular resolution, eq. \ref{oval2theta}), and use the qualitative distances generated in Algorithm \ref{alg-qual-dist} as edge lengths to place the vertices. This process gives us a rough representation which we will further refine in the next stages.

    \subsection{Initial Reconstruction}%
    \label{section-2.1-initial-reconstruction}

    In the current implementation, the input polyline's initial $xy_0$ ($p_0$) is used to help orient and anchor our reconstructed polyline. Tracing the \eopra{m} hull is performed like:

    %!TEX root = ../dce-eopra-report.tex
\noindent
\begin{minipage}{\linewidth}
\begin{algorithm}[H]
\caption{Tracing the Qualitative Hull of an $n$-sided Polygon}\label{reconstruct-trace}
\begin{algorithmic}

    \Statex
    \State $T\gets$ list of $n$ \eopra{m} directions converted back to angles
    \State $D\gets$ list of $n$ \eopra{m} qualitative edge lengths 
    \State $P\gets$ empty list to hold reconstructed points
    \Statex
    \State $P_0\gets$ input polygon's $p_0$%
    \Comment use input polygon's $p_0$ as our $p_0$
    \Statex
        \For{$i\gets 0, \dotsc, n$}
            \State $p_{i+1}\gets$ \textsc{nextXY}($p_i,~D_i,~T_i$)
        \EndFor
    \Statex 
    \State \textbf{return} $P$%
    \Comment return list $P$ of points generated by \texttt{nextXY}

\end{algorithmic}
\end{algorithm}
\end{minipage}
    \vspace{\baselineskip}

    The \texttt{nextXY} method simply returns the next $x,y$ coordinates using the standard vector equations of:
    \begin{equation}
        \begin{aligned}
        x_{n+1} &= x_{n} + \left(distance * \cos{\theta}\right) \\
        y_{n+1} &= y_{n} + \left(distance * \sin{\theta}\right)
        \end{aligned}
    \end{equation}
    
    Once completed, the process returns a sequence of points representing our first attempt at reconstructing the input polygon. Given an input polygon with $n$ points, our initial reconstruction will have $n+1$ points: this is because our reconstructed polygon has two points representing $p_0$. The first of these points is the $p_0$ from the input polygon, and the second is the $p_n$ (the last coordinate) we generated when tracing the hull. In a perfect world, these two values would be identical at this stage, however we often need to adjust the initial reconstruction such that we can achieve closure and remove the duplicate point. 

    To this extent, we begin the refinement process by creating qualitative descriptors of our initial reconstructed polygon just as we did with the original input polygon. This allows us to guide the refinement process by comparing our reconstructions with the input shape via comparing qualitative descriptors. Mainly, we look to the pairwise \eopra{m} direction matrices as measures of similarity.

    \subsection{Refinement}%
    \label{section-2.2-refinement}

    In most cases, the initial reconstruction will produce a polyline which is not closed, and which only matches the qualitative hull descriptors. The goal of the refinement stage is two-fold: 
    \begin{itemize}
        \item achieve closure and 
        \item minimize the difference between the initial polyline's pairwise \eopra{m} direction matrix and the reconstructed polyline's pairwise \eopra{m} direction matrix.
    \end{itemize}

    The refinement process operates by recursively generating and selecting valid ``successor polylines,'' with the two goals above in mind. Successor polylines are created by applying a set of four adjustments to each vertex of the reconstructed polyline (minus the endpoints). These adjustments are: 
    \begin{itemize}
        \item extend outgoing edge length
        \item contract outgoing edge length
        \item increase turn angle, and
        \item decrease turn angle.
    \end{itemize}

    Successor polylines are only passed on to the next steps if: the resulting polyline does not increase either the closure gap or the \eopra{m} direction matrix difference. These checks ensures that only ``good'' (closer to closure) and ``legal'' successors (those which do not break any existing matches between \eopra{m} direction matrices) are returned.

    Using the initial reconstructed polyline, its closure gap, and its pairwise \eopra{m} direction matrix difference from the initial shape as a baseline, the refinement begins by returning all possible valid polylines which are ``one step,'' or one adjustment away from the current state.

    When complete, the method returns a list of valid successor polylines, sorted by their closure gap. The ``best'' of these valid successors (as measured by minimizing the closure gap and \eopra{m} direction matrix difference) is then used to repeat the process. The \texttt{generateSuccessors} method is outlined below.

    %!TEX root = ../dce-eopra-report.tex
\noindent
\begin{minipage}{\linewidth}
\begin{algorithm}[H]
\caption{Generating Successor Polylines}\label{reconstruct-successors}
\begin{algorithmic}
    \Input initial reconstruction $s_0$, goal polyline \eopra{m} direction matrix $goal$
    \Statex
    \State $\Delta_0\gets\textsc{diff}(s_0,goal)$%
    \Comment initial \eopra{m} matrix difference
    \State $gap_0\gets\textsc{ClosureGap}(s_0)$%
    \Comment initial closure gap
    \State $S\gets$ empty list to hold successors
    \State $S_{valid}\gets$ empty list to hold valid successors
    \Statex
    \For{$i\gets 1, 2, \dotsc, n-1$}
        \For{$adjustment,~mutator$ in $adjustments,~mutators$}
            \State{$S_i\gets\textsc{mutator}(p_i,~adjustment)$}%
            \Comment perform $adjustment$ at vertex $p_i$
        \EndFor
    \EndFor
    \Statex
    \For{$i\gets 0, 1, \dotsc, n-1$}
        \If{$\textsc{diff}(S_i,goal) < \Delta_0$}
            \State{$\Delta_0\gets\textsc{diff}(S_i,goal)$}
            \State{$S_{valid_i}\gets S_i$}
        \EndIf
    \EndFor
    \Statex
    \State \textbf{return} $S_{valid}$%
    \Comment return list of valid successor polylines
    \Statex
    \Output list of valid successors ``one-step'' away from $s_0$
\end{algorithmic}
\end{algorithm}
\end{minipage}
    \vspace{\baselineskip}

    Initially, the angular and distance adjustments are fairly coarse. This allows for large or otherwise obvious adjustments to be made as soon as possible. However, if the algorithm determines that either: 
    \begin{itemize}
        \item there are no valid ``successors'' that improve the polyline, or 
        \item the differences are no longer moving towards 0, it will decrease the magnitude of the adjustments.
    \end{itemize}

    The largest angular adjustment is programmatically determined by the \eopra{m} granularity $m$ (one-quarter of the angular resolution), and the minimum angular adjustment is arbitrarily set to 1/20$^{th}$ of the initial adjustment. Given the nature of the edge length mutator, the distance adjustments are defined slightly differently (as percent of current length), but effectively range from from $currentLength \times 2.0$ to $currentLength \times 0.5$.

    A rough outline of the refinement process as a whole is given below.

    %!TEX root = ../dce-eopra-report.tex
\noindent
\begin{minipage}{\linewidth}
\begin{algorithm}[H]
\caption{Refining Successor Polylines}\label{refine-successors}
\begin{algorithmic}
    \Input successor polyline $q_0$, input polyline $p_0$
    \Statex
    \State $\Delta_0\gets$\textsc{diff}($q_0,p_0$)%
    \Comment initial \eopra{m} matrix difference
    \State $gap_0\gets\textsc{ClosureGap}(q_0)$%
    \Comment initial closure gap
    \State $S\gets$\textsc{Successors}($q_0,p_0$)%
    \Comment sorted successor polylines
    \Statex
    \Loop
        \State $gap_{current}=$\textsc{ClosureGap}($S_0$)
        \If{$gap_{current} \le$ minimum appreciable gap}
            \State \textbf{break}%
            \Comment gap is negligible, break
        \EndIf
        \If{$gap_{current} \ge gap_0$}
            \If{scores not improving}
                \State \textbf{break}%
                \Comment no improvements left
            \EndIf
            \If{adjust factor $<20$}
                \State adjust factor $+= 1$%
                \Comment try increasing adjustment resolution
            \EndIf
        \EndIf
        \Statex
        \State $gap_0\gets gap_{current}$%
        \Comment otherwise we have a better gap
        \State $S\gets$\textsc{Successors}($S_0,p_0$)%
        \Comment so use best state to get more successors
    \EndLoop

\end{algorithmic}
\end{algorithm}
\end{minipage}
    \vspace{\baselineskip}
    
    The refinement process will stop when either: 
    \begin{itemize}
        \item the closure gap becomes negligible, or 
        \item the score stops moving even after changing the adjustment resolution \texttt{n} times.
    \end{itemize}

    As this stage, the algorithm will try to ``snap'' the polyline shut by setting the endpoint vertices as equal (as long as doing so does not change any qualitative relations). Snapped or not, the refinement process then returns the final state.

    \section{Preliminary Results}%
    \label{preliminary-results}

    A number of sample polylines with a vertex count ranging from 6 to 15 have been tested with the presented processes. Given an \eopra{m} granularity of $m=8$, all of the sample polylines can successfully be translated to \eopra{8} qualitative descriptions, and reassembled back into approximately similar polylines which achieve both closure and 0-difference \eopra{8} direction hull descriptions, with near-0 \eopra{8} pairwise direction matrix differences. While the refimenent process arbitrarily permits up to 100 iterations and up to 20 adjustment refinements, most of the samples tested reach closure well before those limits.

    Following are three sample results, using input polylines with 7, 9, and 10 vertices. Three shapes are shown for each sample input polyline: from left to right, the input polyline, the initial reconstruction, and the final adjusted polyline. In all of the cases below, the input and final polylines have identical \eopra{8} hull direction descriptions.

    \begin{figure}[!ht]
    
        \begin{center}
        \includegraphics[width=0.8\textwidth]{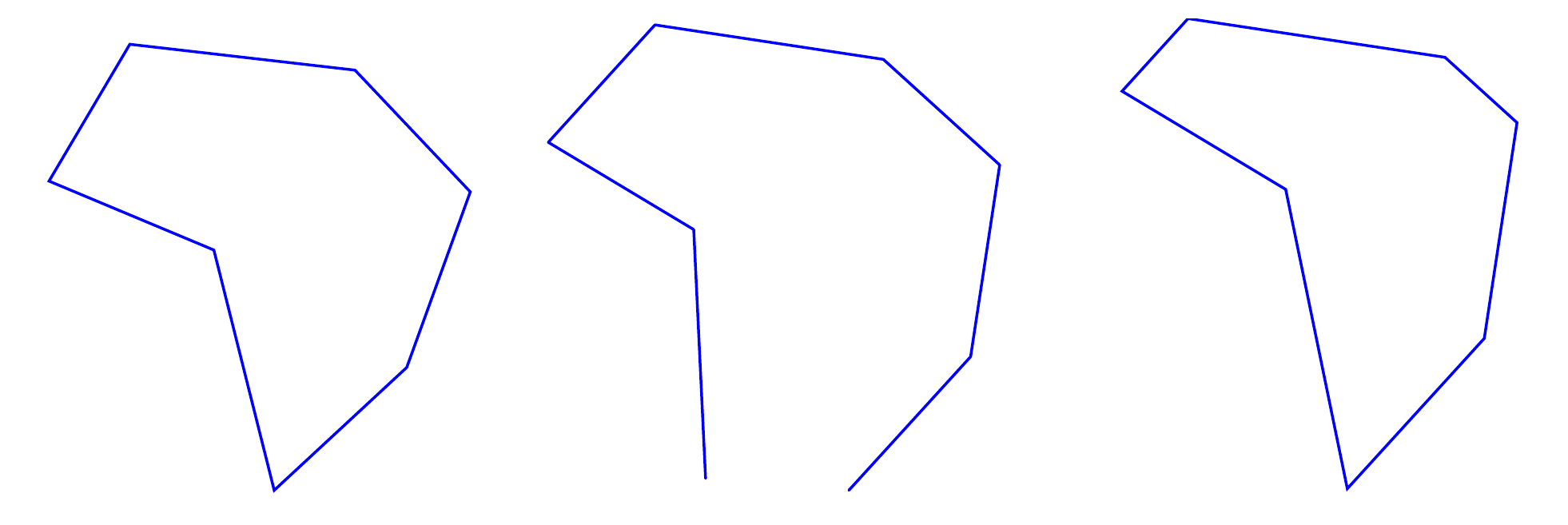}
        \caption{%
            \label{fig:results-africa7}
            7-vertex representation of Africa. Final polyline reached after 3 iterations. Complete (initial, final) \eopra{8} direction matrix differences: $\left(\frac{9}{49},~\frac{5}{49}\right)$.
        }
        \end{center}
    
    \end{figure}

    \begin{figure}[!ht]
    
        \begin{center}
        \includegraphics[width=0.8\textwidth]{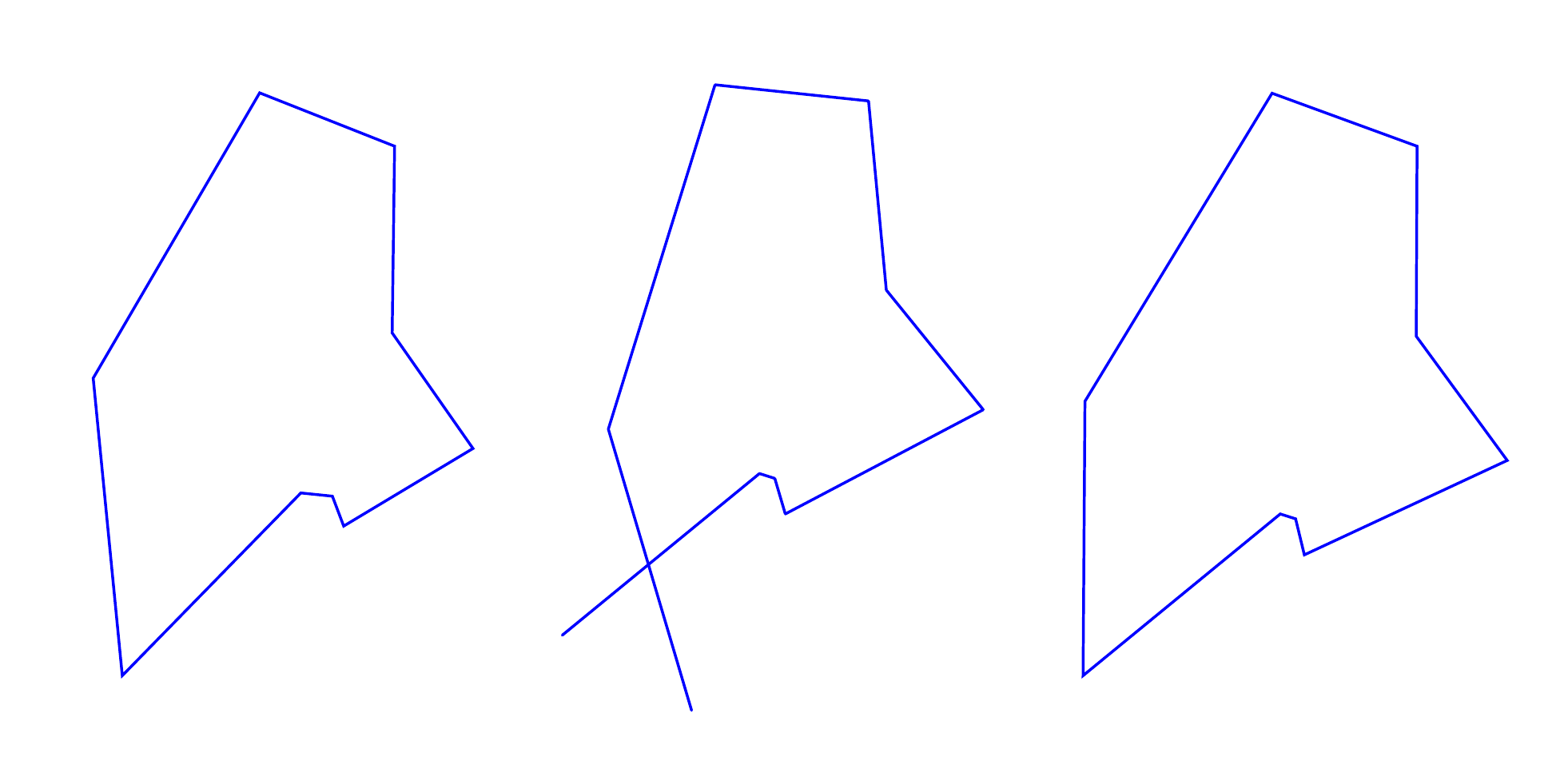}
        \caption{%
            \label{fig:results-maine9}
            9-vertex representation of the state of Maine. Final polyline reached after 6 iterations. Complete (initial, final) \eopra{8} direction matrix differences: $\left(\frac{21}{81},~\frac{7}{81}\right)$.
        }
        \end{center}
    
    \end{figure}

    \begin{figure}[!ht]
    
        \begin{center}
        \includegraphics[width=0.8\textwidth]{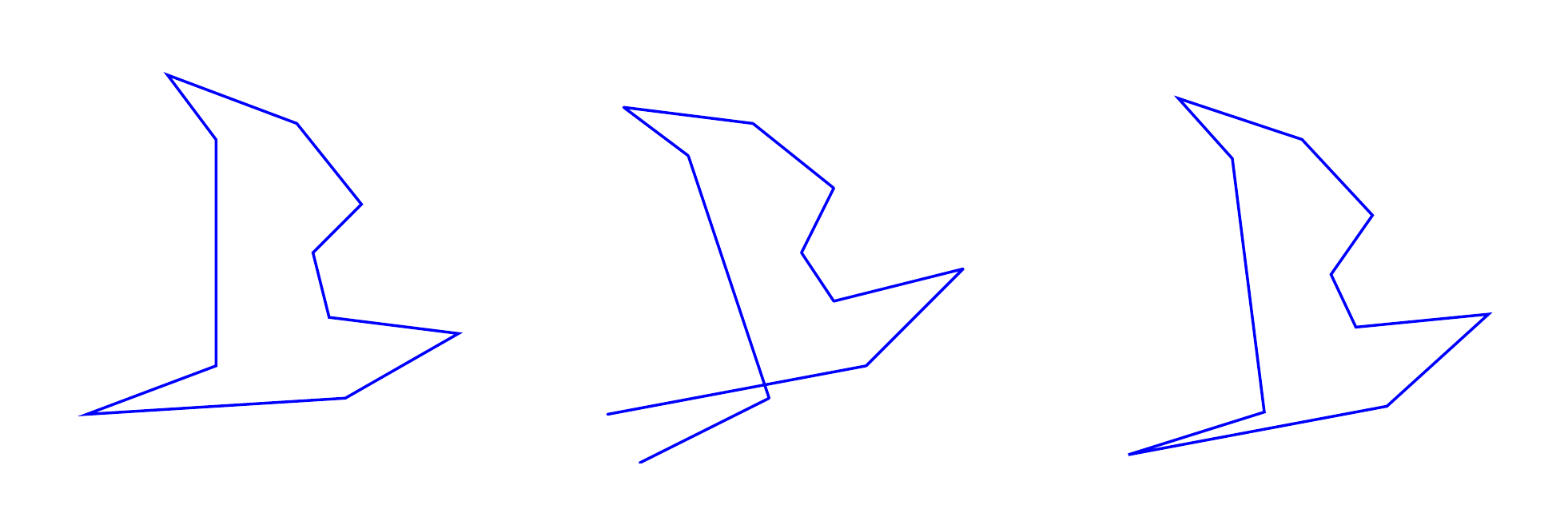}
        \caption{%
            \label{fig:results-eagle10}
            10-vertex representation of a bird. Final polyline reached after 5 iterations. Complete (initial, final) \eopra{8} direction matrix differences: $\left(\frac{18}{100},~\frac{9}{100}\right)$. 
        }
        \end{center}
    
    \end{figure}

    \section{Conclusion and Outlook}%
    \label{discussion-futwork}

    We have developed a qualitative shape description schema based on the qualitative relative direction and distance calculus \eopra{m}. With this new method, we implicitly have a schema for similarity of shapes which has the potential to be cognitively adequate. To enable tests for this cognitive adequacy, the qualitative shape representation must be reversible. With our approach, we can take the qualitative shape representation and generate prototypical specific shapes that match the abstract description. Our work as described in this report is the very first QSR-based shape description capable of generating prototypical shapes based on the abstract qsr-based representation. Results presented in this report are produced by a small set of Python programs developed to perform the deconstruction and reconstruction tasks.

    Future work includes performing empirical studies with human test subjects to test the cognitive adequacy of our new qualitative shape representation. We also will investigate the application of our shape representation to the formalization of affordances. Affordances are the perceived potential function of everyday objects like chairs, desks, stairs, tables etc. These affordances can assist in the task of object categorization based on 3D shape information \cite{raubal2008functional, wunstel2004automatic, isli2000topological, moratz2008affordance, hildebrandt1999, moratz1997visuelle, moratz1994controlling}.

    \subsection*{Acknowledgement}

The authors would like to thank Jan Oliver Wallgr{\"u}n for helpful discussions related to the topic of this paper.
Our work was supported by 
the National
Science Foundation under Grant No. CDI-1028895.

\end{document}